\documentclass[11pt,twocolumn,twoside]{article}
\usepackage{fully3d}
\usepackage{amssymb}
\usepackage{amsmath}

\begin{document}

%-------------------------------------------------------------------------------------------
%%%%% add your title here %%%%%
\title{Image Synthesis for Data Augmentation in Medical CT using Deep Reinforcement Learning} 

%%%%% add authors and affiliations here %%%%%
\author[1]{Arjun Krishna}
\author[1]{Kedar Bartake}
\author[2]{Chuang Niu}
\author[2]{Ge Wang}
\author[3]{Youfang Lai}
\author[3]{Xun Jia}
\author[1]{Klaus Mueller}

\affil[1]{Computer Science Department,
          Stony Brook University, Stony Brook, NY USA}

\affil[2]{Department of Biomedical Engineering, Center for Biotechnology \& Interdisciplinary Studies, Rensselaer Polytechnic Institute, Troy, NY USA}

\affil[3]{Department of Radiology Oncology,
          UT Southwestern Medical Center, Dallas, TX USA}
          
%%%%% don't change these 2 lines %%%%%
\maketitle
\thispagestyle{fancy}

%-------------------------------------------------------------------------------------------
%%%%% add your summary (abstract) here               %%%%%%
%%%%% use footnotesize for this section              %%%%%%
%%%%% please stick to the customabstract environment %%%%%% 

\begin{customabstract}
Deep learning has shown great promise for CT image reconstruction, in particular to enable low dose imaging and integrated diagnostics. These merits, however, stand at great odds with the low availability of diverse image data which are needed to train these neural networks. We propose to overcome this bottleneck via a deep reinforcement learning (DRL) approach that is integrated with a style-transfer (ST) methodology, where the DRL generates the anatomical shapes and the ST synthesizes the texture detail. We show that our method bears high promise for generating novel and anatomically accurate high resolution CT images at large and diverse quantities. Our approach is specifically designed to work with even small image datasets which is desirable given the often low amount of image data many researchers have available to them.  
\end{customabstract}

%-------------------------------------------------------------------------------------------
%%%%% main text                                                %%%%%    
%%%%% remove the dummy content and put your own content here   %%%%% 
%%%%% feel free to choose your own section titles              %%%%% 
%%%%% you don't need to put the content in a separate tex file %%%%%

% dummy_content.tex shows how to add sections, figures, tables, formulas, and references
% remove the following line, it just adds dummy content
\section{Introduction}
One of the key challenges in unlocking the full potential of machine and deep learning in radiology is the low availability of training datasets with high resolution images. This scarcity in image data persists predominantly because of privacy and ownership concerns. Likewise, publicly available annotated high resolution image datasets are also often extremely small due to the high cost and small number of human experts who have the required amount of medical knowledge to undertake the labeling task. With insufficient data available for model training comes the inability of these networks to learn the fine nuances of the space of possible CT images, leading to the possible suppression of important diagnostic features and in the worst case making these deep learning systems vulnerable to adversarial attacks. We present an approach that can fill this void; it can synthesize a large number of novel and diverse images using training samples collected from only a small number of patients. 

%Our method extends our previous work \cite{krishna2019medical} 
%our style transfer framework to generate  anatomically correct and highly diverse full resolution images from an extremely small training dataset.

%\subsection{Use of This Document}
Our method is inspired by the recent successes of Deep Reinforcement Learning (DRL)~\cite{li2017deep,franccois2018introduction} in the game environments of Atari~\cite{mnih2013playing}, Go and Chess~\cite{silver2018general} which all require the exploration of high-dimensional configuration spaces to form a competitive strategy from a given move. It turns out that this is not too different from generating plausible anatomical shapes in medical CT images. Our methodology combines the exploratory power of Deep Q Networks \cite{bworld} to optimize the parameter search of geometrically defined anatomical organ shapes, guided by medical experts via quick accept and reject gestures. This need for feedback eventually vanishes, as the network learns to distinguish valid from invalid CT images.
%We conduct the opportunistic search  in a configuration space generated via Principal Component Analysis (PCA) from the existing image data. 

During the generation, once the anatomical shapes for a novel CT image have been obtained from  the DRL module, we use a style transfer module, designed for the texture learning of component organs and tissues~\cite{krishna2019medical}, to generate the corresponding high resolution full-sized CT image. 
% USE SOMEWHERE ELSE Since DRL networks learn by optimizing on results (rewards) that derive from the rules of the environment (mostly games); we have to design our own environment to stimulate the learning of an effective strategy for exploring the anatomical shape space for facilitating diversified yet accurate image generation. Our paper describes setup of one such environment.
%Nonetheless, we were still motivated by their success in the style transfer based approach and decided to use these perceptual losses for the anatomical accuracy of the appearance of every organ for the generated full-sized CT image. 
To the best of our knowledge, our proposed approach is the first attempt to incorporate DRL networks for the synthesis of new diverse full-sized CT images.
%In Section 2 we describe our method at greater detail. Section 3 shows some results and Section 4 ends with conclusions. 
 
 %The next section, Section II, describes the cGAN architecture in combination with the perceptual losses that we have used. The following section, Section III, presents first experimental results we have obtained with our prototype for visual analysis. The results confirm the great promise these types of losses have in image synthesis for CT. We end the paper with a discussion and pointers to future work.

\section{Methods}

We adopt a two-step approach for synthesizing the full-resolution CT images. The first step consists of creating an anatomically accurate semantic mask (SM) for the image; this is the focus of this paper's discussion. The second step uses our existing style transfer network~\cite{krishna2019medical} to render anatomically accurate texture into the different portions of the generated SM.  

As shown in Figure 1 (next page), step 1 consists of two phases. The first phase includes data pre-processing and training of a classifier following a traditional Convolutional Neural Network architecture \cite{Goodfellow-et-al-2016} for classifying images. The data pre-processing stage produces the SMs of the high-resolution CT training images; it represents the annotated segmentations of the various anatomical features, such as organs and skeletal structures, as a set of 2D curves which are then geometrically parameterized as B-splines of order n for n+1 control points {\{(x$_{i}$, y$_{i}$)\}}$_{i=1}^{n}$. The control points of the anatomical features are stored as sequences of coordinates into vectors and then embedded into a lower dimensional space obtained via PCA. PCA is attractive since it preserves the spatial relationships of the SMs, has a linear inverse transform, and identifies a reduced orthogonal basis that approximates the shape of the SM statistical distribution well. Next, to train the classifier sufficiently, we generate a large number (on the order of 10,000) new semantic masks by interpolating in this PCA space and group these images into clusters via k-means. The clusters are then manually labeled by experts as good and bad image sets and the classifier is then trained on these clusters. The classifier thus represents an approximation of control points that could serve as valid semantic masks.

\begin{figure*}
  \centering
  \includegraphics[height=7.0cm]{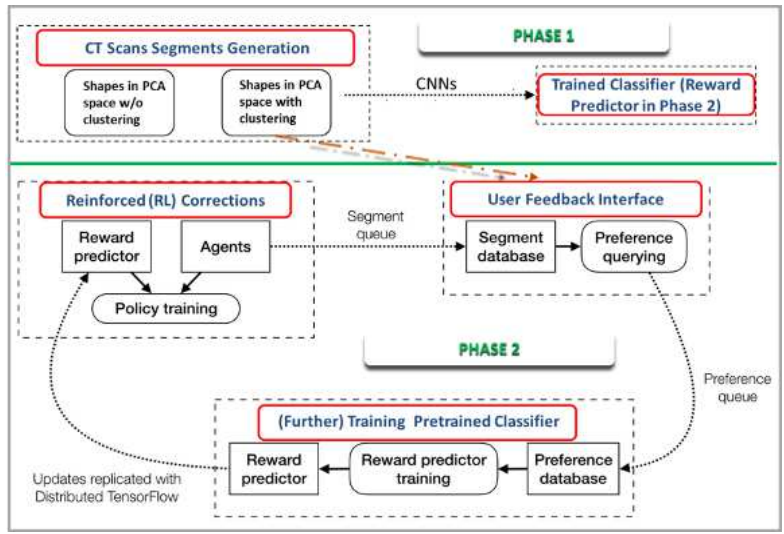}
  \caption{Two-Phase box diagram for training RL agents. The pre-trained classifier in Phase 1 is used as reward predictor in Phase 2. Segment refers to the resulting SM from agents' actions. Preference refers to the user preference of one segment (SM) over other.}
  \label{fig:dummyfigure}
\end{figure*}

Phase 2 uses this trained classifier as the reward predictor in our Reinforcement Learning Environment (RLE). DRL networks learn by optimizing on results via a reward mechanism that derives from the rules of the environment. This environment serves to stimulate the learning of an effective strategy for exploring the anatomical shape space to facilitate a diversified yet accurate image generation. Our specific environment for DRL involves a user-feedback interface that consists of a front-end where linear interpolations between the semantic masks of two distinct valid SMs are corrected by the agents of the RLE followed by the expert user marking them as good or not. This feedback is then used to further train the classifier/reward predictor such that it can give better predictions of the actual rewards to the agents as they try to correct future interpolations. Hence the agents in RLE and the reward predictor are trained asynchronously. As the reward predictor gets better, so do the actions of the agents and consequently we gain more semantic masks representing valid plausible anatomy. 
%that lie in between the two random patients.
%As the dataset gets bigger, perfectly generated interpolations could in turn be added to the training dataset in accordance to the user feedback to generate even more interpolations between patients. The resulting trained agents could then be used to correct any semantic mask generated in the PCA space since they have learned a good approximation of the true high-dimensional space in which the control points of valid (anatomically accurate) semantic masks lie in. 

Our contributions are as follows:
\begin{itemize}
    \item We discuss a robust way of learning anatomical shapes via their geometrical representations of B-splines and their interpolations / samplings in PCA space.
    \item We define an environment where the true image space of the anatomical shapes could be discovered without the supporting dataset via Reinforcement Learning.
    \item We build a visual user-interface where users can control and guide the generation process. Once sufficiently trained, users have the option to add the generated images to the training dataset.
\end{itemize}

\begin{figure*}
  \centering
  \includegraphics[height=5 cm]{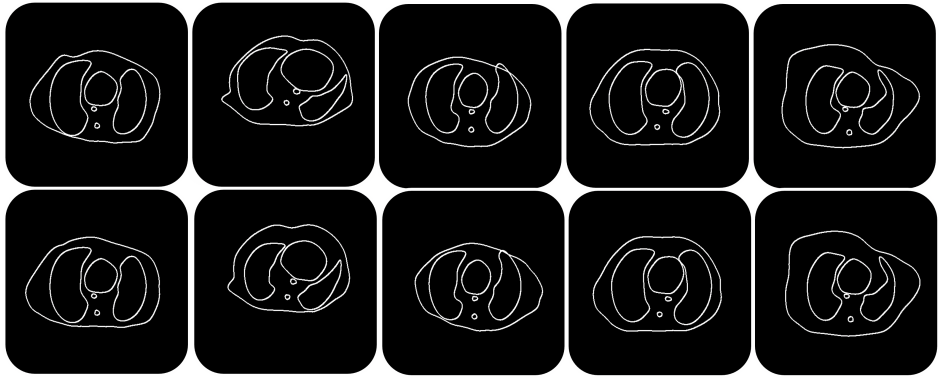}
  \caption{The first row shows linearly interpolated SMs for a lung CT image. The second row shows their improved counterparts from RL agents. In the first three columns, the agents tries to make them more symmetric and remove intersections. For anatomically accurate interpolated SMs, agents don't make much change as seen in the fourth column. The fifth column represents the anatomical space in our PCA for which agents have not yet been trained on and would improve with incoming user feedback}
  \label{fig:dummyfigure}
\end{figure*}

\subsection{General Interpolation Framework: B-Splines and PCA Interpolation}
Curvature is a central morphological feature of organs, tissues, cells, and sub-cellular structures \cite{mary2019kappa}. Hence we represent the curve shapes by the set of control points with strongest curvatures between some predefined distances across the whole curves depicting organs, skeletal structures, etc., we shall refer to it as \textit{anatomical shapes}. These control points also integrate easily with B-spline curves to decode them back into full curves. B-spline curves provide flexibility to represent these anatomical curves~\cite{wenckebach2005capturing} since the degree of a B-spline curve is separated from the number of control points. Hence lower degree B-spline curves can still maintain a large number of control points and the position of a control point would not change the shape of the whole curve (local modification property). Since B-splines are locally adjustable and can model complex shapes with a small number of defined points, they are an excellent choice to model anatomical shapes with control points selected based on strong curvatures. 

Since each semantic mask (SM) is expressed as a set of control points, we embed the training data SMs in a lower dimensional space via Principal Component Analysis (PCA). %PCA is attractive since it preserves the spatial relationships of the SMs, has a linear inverse transform, and identifies a reduced orthogonal basis that approximates the shape of the SM statistical distribution well. 
The PCA model is used to reconstruct the anatomical shapes of the training dataset giving us a repository of coefficients for eigen-vectors that make plausible anatomy for lung CT SMs. We can then reconstruct new anatomy curves by sampling these coefficients. Each type of anatomical shape, such as left lung, right lung, torso, spinal cord, esophagus, and heart, forms a dedicated subspace of SM vectors and is represented as a multivariate Gaussian with mean (for each coefficient of the corresponding eigen-vector) and co-variance matrix. The set of anatomical shapes for a specific SM are interlinked so they can be jointly used in the interpolation procedure.
%A clustering algorithm such as k-means is used in generating SMs with the correlated shapes with a plausible anatomy. 
%We use the KNN (K nearest neighbors) mechanism to look up the nearest neighbors of a sampled anatomical shape. These nearest neighbours then give a set of linked coefficients for every other anatomical shape's  which can then be sampled using their multivariate Gaussian distributions.
In our initial implementation we represented all anatomical shapes of the training SMs as a single vector to form a single multivariate Gaussian. In practice, however. this approach does not work well and fails to generate SMs with correlated anatomical shapes.

One way to generate a novel SM is to take any two available SMs and linearly interpolate between the two. One problem with this approach is that with small training datasets there is not enough variety to construct an accurate PCA decomposition. leading to
%the eigen decomposition among eigen vectors would
noise and subsequently to erroneous features in the generated SM. Also, accurate anatomical shapes do not occupy a perfectly linear space even in heavily reduced dimensions and the interpolation on the eigen-vectors still limits the number of novel anatomical shapes that can be generated since the set of images between which the interpolation is being done is small. To overcome these limitations, we introduce the powerful mechanism of DRLs within our environment which we describe in the next section.

\subsection{User Assisted Deep Reinforcement Learning}
%As mentioned, the small dataset would severely affect the ability of our (PCA) model to explore the hidden valid spaces in the PCA space which represent plausible anatomies not present in the training data. 
We propose to solve the aforementioned problem with PCA space exploration using Deep Reinforcement Learning, obtaining user feedback via a dedicated user interface. We ask a user to interpolate between two generated anatomies by moving a slider. We then present small perturbations made by the agents in the Deep Q Learning environment to the linear interpolation and present these to the user as alternative results. The user picks which ones are better and which ones are worse and submits his or her feedback via the interface. The submitted preferences train a CNN (Convolutional Neural Network) based image classifier that is simultaneously used as a reward predictor for training the agents in the Deep-Q Learning algorithm. Our approach of using a reward predictor to predict rewards based on user feedback mainly borrows from the work of Christiano et al.\cite{christiano2017deep} who utilize user feedback on video clips of game play to train a reward predictor. 

As shown in Figure 1, we pre-train the reward predictor during the data processing stage. By modifying the parameters in the clustering (via k-means), we can visibly alter the quality and anatomical accuracy of the generated SMs when interpolating in PCA space. These groups of SMs can be used to pre-train the reward predictor that is used in our DRL environment where it is further fine-tuned with the help of user feedback. The trained reward predictor on submitted user preferences then help the agents in learning the perturbations that need to be applied to the coefficients of eigen-vectors representing a SM while interpolating in between any two random SMs. Note that because of this setup once agents are trained, they can also be used to "fix" any generated SM interpolated on the PCA space. With the help of user verification, we add perfectly generated SMs in the training dataset that are then used to interpolate more novel SMs hence expanding the known PCA space representing valid anatomy. This helps our SM generating interface get better with the usage by the users.

\begin{figure*}
  \centering
  \includegraphics[height=3.9 cm]{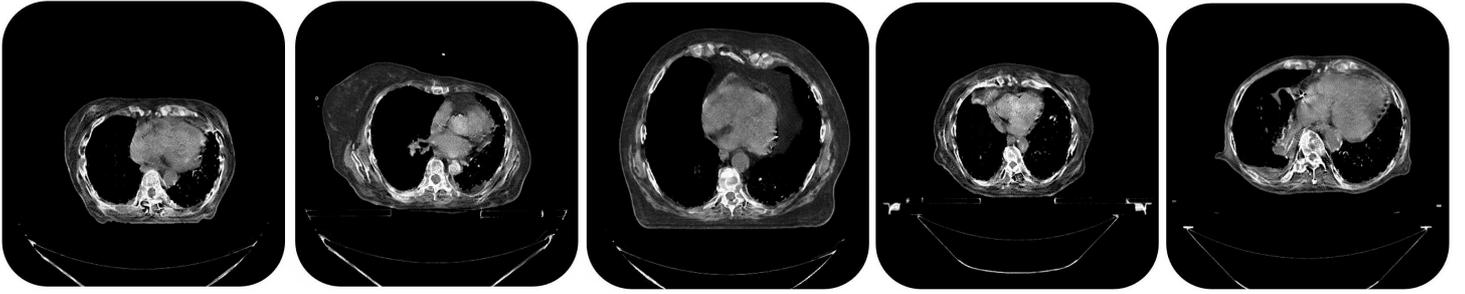}
  \caption{Some stylized CT images, generated by linear SM pair interpolation, and corrected with the RL framework.}
  \label{fig:dummyfigure}
\end{figure*}

\subsection{Loss Function, Input/Output and Network Architecture of Deep-Q Agents}
We follow the Deep-Q DRL algorithm used by the authors of Atari \cite{mnih2013playing}. We maintain a policy $\pi$ that takes the observation state O as input and gives an action A to be performed; $\pi$ : O ${\xrightarrow{}}$ A. The reward predictor takes the resulting image as input and gives a reward estimate R; $\hat{r}$ : O x A ${\xrightarrow{}}$ R. For training our policy $\pi$ we use the traditional Deep-Q loss:
\begin{align}
y_{i} = \mathbb{E}_{s'\sim \varepsilon}[\hat{r} + \gamma max_{a'}Q(s',a';\theta_{i-1}))^{2}]
\end{align}
\begin{align}
L_{i} (\theta_{i}) = \mathbb{E}_{s,a \sim \rho(\cdot)} [(y_{i} - Q(s,a;\theta_{i}))^{2}]
\end{align}
where $y_{i}$ represents the discounted reward estimate from iteration i and $\rho$(s,a) represents the distribution of all states and actions applicable on those states. Since our states are sequences of coefficients for representing the control points of every organ (thereby representing the set of anatomical shapes constituting SMs), we use a neural network using six fully connected layers to estimate the second term; $Q(s,a;\theta_{i})$ in equation (2). The parameters from the previous iteration $\theta_{i-1}$ are held fixed when optimising the loss function $L_{i}(\theta_{i})$ and are estimated via stochastic gradient descent.

\subsection{Loss Function, Input/Output and Network Architecture of Reward Predictor}
Once the agents modify the contributions of the eigen-components, the resulting anatomical shapes are assembled into a SM and sent to a six layer CNN with batch normalization layers and relu activations \cite{Goodfellow-et-al-2016}. The CNN classifies the SM image in one of five or six categories indicative of their anatomical accuracy according to which a reward is assigned to the action of agent. The policy $\pi$ interacts with the environment to produce a set of trajectories ${\{\tau^{1}...\tau^{i}\}}$. A pair of such trajectory results (SMs) are selected and are sent to our front-end for user feedback. To fine-tune the reward predictor further we use the cross entropy loss between the predictions of the reward predictor and user feedback $\nu$ \cite{christiano2017deep}.
\begin{align}
loss(\hat{r}) = \sum_{\tau^{1},\tau^{2},\nu} \nu(1)log\hat{P}[\tau^{1} \succ \tau^{2}] + \nu(2)log\hat{P}[\tau^{2} \succ \tau^{1}]
\end{align}
where under the assumption that user's probability of preferring a SM over other should depend exponentially on the true total reward over the SM's trajectory; $\hat{P}[\tau^{1} \succ \tau^{2}]$ could be expressed as:
\begin{align}
\hat{P}[\tau^{1} \succ \tau^{2}] = \frac{exp\sum\hat{r}(s_{t}^{1},a_{t}^{1})}{exp\sum\hat{r}(s_{t}^{1},a_{t}^{1})+exp\sum\hat{r}(s_{t}^{2},a_{t}^{2})}
\end{align}
As evident from figure 1, the above two networks are trained asynchronously. With increasing data from the user's feedback, the reward predictor gets better which helps better train the RL agents.

\bigskip

\section{Results, Future Work and Conclusion}

Figure 2 shows corrected SMs via RL agents from badly formed counterparts which were interpolated linearly between two generated SM images. In most cases, our RL agents are able to correct the obvious errors like the intersections between the organ curves or the sharp unnatural bends in the boundaries of torsos, but as evident from the example in the last column of the figure, for some badly formed SMs the agents are unable to make better SMs. That's because we need more user feedback for training the reward predictor enough to make agents respond to a wide range of generated SMs. With more feedback that the reward predictor would receive, the agents could be trained better for responding to the generated SMs. Figure 3 shows stylized CT images on corrected SMs. 

For future work, we intend to modify the user-interface to enable faster user interaction hence enabling larger feedback collection quickly for more efficient training of the reward predictor and the RL agents. We also plan to make the texture learning more robust on varied SMs and not just lung CT SMs. We also intend to extend our framework for learning and generating pathology which should integrate well with our two step approach. At the current time, we generate volumes slice by slice. For better continuity across slices, we plan to learn anatomical curves directly in 3D volumes, using B-spline patches.

%-------------------------------------------------------------------------------------------
\printbibliography

\end{document}